\documentclass[conference]{IEEEtran}
\IEEEoverridecommandlockouts

\usepackage{cite}
\usepackage{amsmath,amssymb,amsfonts}
\usepackage{algorithmic}
\usepackage{graphicx}
\usepackage{textcomp}
\usepackage{microtype}
\usepackage{hyperref}
\usepackage[most]{tcolorbox}
\usepackage{url}
\usepackage{booktabs}
\usepackage{multicol}
\usepackage{colortbl}
\usepackage{subfigure}
\usepackage{epsfig,color}
\usepackage{array}
\usepackage{lipsum}
\usepackage{amsmath}
\usepackage{amssymb}
\usepackage{float}
\usepackage{pifont}
\usepackage{makecell}  
\usepackage{multirow}
\usepackage{tikz}
\usepackage{hyperref}
\usepackage{makecell}
\usepackage{CJKutf8}
\usepackage{tikz}
\usepackage[edges]{forest}
\definecolor{hidden-draw}{RGB}{20,68,106}
\definecolor{hidden-pink}{RGB}{255,245,247}
\newcommand{\cmark}{\text{\ding{51}}}
\newcommand{\xmark}{\text{\ding{55}}}

\def\BibTeX{{\rm B\kern-.05em{\sc i\kern-.025em b}\kern-.08em
    T\kern-.1667em\lower.7ex\hbox{E}\kern-.125emX}}

\newif\ifsubmission
\submissionfalse

\ifsubmission
\newcommand{\faraz}[1]{}
\newcommand{\anwar}[1]{}
\newcommand{\azal}[1]{}

\else
\newcommand{\faraz}[1]{\textit{\textcolor{red}{{Faraz:}#1}}}

\newcommand{\azal}[1]{\textit{\textcolor{blue}{{Azal:}#1}}}

\fi

\begin{document}

\title{Personalized Federated Learning Techniques: Empirical Analysis}

\author{
    Azal Ahmad Khan\textsuperscript{1}, Ahmad Faraz Khan\textsuperscript{2}, Haidar Ali\textsuperscript{2}, and Ali Anwar\textsuperscript{1} \\
    \textsuperscript{1}Department of Computer Science and Engineering, University of Minnesota-Twin Cities \\
    \textsuperscript{2}Department of Computer Science, Virginia Tech \\
    Email: \{khan1069, aanwar\}@umn.edu, \{ahmadfk, haiderali\}@vt.edu
}


\maketitle

\begin{abstract}
Personalized Federated Learning (pFL) holds immense promise for tailoring machine learning models to individual users while preserving data privacy. However, achieving optimal performance in pFL often requires a careful balancing act between memory overhead costs and model accuracy. This paper delves into the trade-offs inherent in pFL, offering valuable insights for selecting the right algorithms for diverse real-world scenarios. We empirically evaluate ten prominent pFL techniques across various datasets and data splits, uncovering significant differences in their performance. Our study reveals interesting insights into how pFL methods that utilize personalized (local) aggregation exhibit the fastest convergence due to their efficiency in communication and computation. Conversely, fine-tuning methods face limitations in handling data heterogeneity and potential adversarial attacks while multi-objective learning methods achieve higher accuracy at the cost of additional training and resource consumption. Our study emphasizes the critical role of communication efficiency in scaling pFL, demonstrating how it can significantly affect resource usage in real-world deployments.
\end{abstract}


\section{Introduction}
\label{sec:Introduction}

\textbf{Background.} Google proposed Federated Learning (FL) to aggregate distributed intelligence without compromising data privacy and security ~\cite{konevcny2016federated}. FL is a distributed ML approach that enables training on a large corpus of decentralized data residing on devices like mobile phones, addressing data privacy, ownership, and locality issues and offering advantages over centralized ML ~\cite{bonawitz2019towards}. FL provides benefits such as data security by storing training datasets on devices, improved data diversity by enabling access to heterogeneous data, and increased hardware efficiency by eliminating the need for a single complex central server. FL can be used to build models in user behavior from smartphone data without leaking personal data, such as next-word prediction ~\cite{hard2018federated}, face detection ~\cite{aggarwal2021fedface} ~\cite{bai2021federated} ~\cite{niu2022federated}, voice recognition ~\cite{guliani2021training}, and the healthcare sector~\cite{hard2018federated}, \cite{rieke2020future, antunes2022federated, teo2024federated}.

\textbf{Purpose of personalization.}
Statistical heterogeneity in FL can lead to non-iid (non-independent and identically distributed) data, which refers to a situation where the data distribution among different clients is not the same ~\cite{li2020federated}. Non-iid data can pose significant challenges for machine learning algorithms since they are typically designed for independent and identically distributed data. With non-iid data, traditional ML training algorithms may perform poorly and result in suboptimal models. 
FL applications generally face non-i.i.d and unbalanced data available to devices, making it challenging to ensure good performance across different devices with an FL-trained global model ~\cite{jiang2019improving}.  
The general FL approach faces several fundamental challenges (i) poor convergence on highly heterogeneous data and (ii) lack of solution personalization, (iii) the final model is less appealing for clients, and (iv) limited computation resources ~\cite{tan2022towards, fallah2020personalized}. These issues deteriorate the performance of the global FL model on individual clients. These challenges can be solved via personalization by allowing for the training of multiple personalized models that are customized to the data and preferences of individual clients. Personalization allows for better performance on highly heterogeneous data.

\textbf{Purpose of the study.}
Personalized Federated Learning (pFL) has been extensively explored in various studies, such as those conducted by Kulkarni et al. ~\cite{kulkarni2020survey} and Zhu et al. ~\cite{zhu2021federated}. These studies typically categorize methodologies based on the pFL techniques. Some studies such as Divi et al. ~\cite{divi2021new} propose metrics for fairness evaluation in pFL, going beyond traditional model performance metrics. However, there is a lack of a standard benchmark that could be used for comparative analysis of pFL techniques based on various real-world factors such as resource consumption, efficiency, and performance. Therefore, our contributions go beyond existing studies on pFL algorithms by offering a comprehensive assessment of various aspects, including performance, per-round time and convergence time, as well as memory overhead. This difference in contributions between our study and extant surveys in pFL is shown in Table~\ref{tab:ExistingSurveys}. Our empirical analysis reveals notable trends in the performance of different pFL categories. For instance, methods that utilize personalized (local) aggregation demonstrate the fastest performance due to their ability to reduce communication and computation requirements per round. Methods that use multi-task learning or transfer learning either by finetuning the global model or learning additional personalized models are much more memory intensive.  We then provided a rationale for these findings, which we believe will be valuable for the development of new pFL algorithms. It is crucial for industry to gain a comprehensive understanding of which algorithms are best suited to their practical applications.

\begin{table*}
\centering
\rowcolors{2}{gray!20}{white}
\begin{tabular}{cccc}
\hline
\rowcolor{purple!10}
\textbf{Paper} & \textbf{Categorization} & \textbf{Empirical Analysis} & \textbf{Evaluation Metrics} \\
\midrule
Kulkarni et al. (2020)~\cite{kulkarni2020survey} & \cmark & \xmark & - \\
Zhu et al. (2021) ~\cite{zhu2021federated}& \cmark & \xmark & - \\
Divi et al. (2021) ~\cite{divi2021new} & \cmark & \cmark & Performance\\
Tan et al. (2022)~\cite{tan2022towards} & \cmark & \xmark & - \\
\midrule
\rowcolor{yellow!20}
\textbf{Ours} & \cmark & \cmark & Performance, Memory Overhead, Per-round \& Convergence Time\\
\hline
\end{tabular}
\caption{Comparison with Existing Surveys}
\label{tab:ExistingSurveys}
\vspace{-0.6cm}
\end{table*}


\textbf{Contributions.} Our contributions are as follows:
\begin{itemize}
\item \textbf{(C1)} The paper encompasses a comprehensive \textbf{empirical analysis of ten pFL techniques}, systematically evaluating their performance across multiple datasets and data splits. These algorithms were compared with two popular baselines FedAvg~\cite{mcmahan2017communication} and FedProx~\cite{li2018convergence}.
\item \textbf{(C2)} Our analysis investigated key aspects, including \textbf{convergence speed, memory overhead, and model performance}, providing a comprehensive view of evaluated algorithms.
\item \textbf{(C3)} Our study analyzes the \textbf{trade-off between memory overhead vs model performance and memory overhead vs convergence time} in pFL. Among other insights from this study, we observed that methods incorporating additional personalized models tend to achieve higher accuracy at the cost of slower performance. 


\end{itemize}



\section{General Federated Learning}
\label{sec:GeneralFL}

\subsection{FedAvg}
\label{fedavg}
FedAvg \cite{mcmahan2017communication} works by iteratively averaging model updates from multiple clients to train a global model in an FL setting. In each round of training, a subset of clients are randomly selected to participate. The current global model is sent to each selected client, which trains the model on their local data using Stochastic Gradient Descent (SGD) and returns the updated model parameters. The central server then aggregates these model updates by taking a weighted average, considering the number of data points at each client, to create a new global model. This process is repeated until convergence. However, FedAvg faces challenges when the client data is heterogeneous. This is a core challenge of FL, especially when data is non-IID, as is often the case in real-world scenarios. Since the model is trained on different data sources, the global model may struggle to generalize well across all clients. This motivates the development of pFL techniques, which aim to address the limitations of traditional FedAvg by adapting models to individual client data.

\subsection{FedProx}
\label{fedprox}
FedProx~\cite{li2018convergence} is an extension of the FedAvg algorithm that introduces a proximal term to the objective function of the local subproblem in FL. This proximal term acts as a regularization term, penalizing large deviations of the local model from the global model. The FedProx algorithm begins with the initialization of a global model. A subset of clients is then selected for each round of training. The selected clients train their local models on their respective local data using stochastic gradient descent (SGD) while minimizing the modified objective function that includes the proximal.
\begin{equation}
\min_{w} h_k(w; w^t) = F_k(w) + \frac{\mu}{2} ||w - w^t||^2.
\label{eq:ditto_loss}
\end{equation}
The local models are aggregated at the central server by taking a weighted average, and the proximal term($\mu$) is updated based on the difference between the local and global models, as shown in Equation~\ref{eq:ditto_loss}. These steps are repeated until convergence.

FedProx offers several advantages in FL. It improves convergence compared to FedAvg, particularly in heterogeneous client data scenarios. The proximal term helps to stabilize the learning process and mitigates the impact of variable local updates, resulting in faster and more robust convergence. Additionally, FedProx has been shown to achieve higher test accuracy than FedAvg on realistic federated datasets, indicating improved model performance. However, while FedProx helps mitigate system and data heterogeneity's impact, finding the right balance through hyperparameter tuning becomes crucial. Setting the hyperparameter $\mu$ appropriately is necessary to ensure it is effective in regularizing the model without overly constraining the fitting to local data. Achieving optimal hyperparameter settings can be time-consuming and requires expertise.
\section{Personalized Federated Learning Algorithms}
\label{sec:PersonalizedFL}

\tikzstyle{my-box}=[
    rectangle,
    draw=hidden-draw,
    rounded corners,
    text opacity=1,
    minimum height=1.5em,
    minimum width=5em,
    inner sep=2pt,
    align=center,
    fill opacity=.5,
    line width=0.8pt,
]
\tikzstyle{leaf}=[my-box, minimum height=1.5em,
    fill=pink!40, text=black, align=left,font=\normalsize,
    inner xsep=2pt,
    inner ysep=4pt,
    line width=0.8pt,
]
\begin{figure*}[t!]
    \centering
    \resizebox{\textwidth}{!}{
        \begin{forest}
            forked edges,
            for tree={
                grow=east,
                reversed=true,
                anchor=base west,
                parent anchor=east,
                child anchor=west,
                base=center,
                font=\large,
                rectangle,
                draw=hidden-draw,
                rounded corners,
                align=left,
                text centered,
                minimum width=4em,
                edge+={darkgray, line width=1pt},
                s sep=3pt,
                inner xsep=2pt,
                inner ysep=3pt,
                line width=0.8pt,
                ver/.style={rotate=90, child anchor=north, parent anchor=south, anchor=center},
            },
            where level=1{text width=9em,font=\normalsize,}{},
            where level=2{text width=20em,font=\normalsize,}{},
            where level=3{text width=11em,font=\normalsize,}{},
            where level=4{text width=7em,font=\normalsize,}{},
            [
                Federated Learning \\ \hspace{1.5em}Algorithms, ver
                [
                    General FL (\S \ref{sec:GeneralFL}) 
                    [
                        FedAvg~\cite{mcmahan2017communication}{, } FedProx~\cite{li2018convergence},
                        fill=orange!20!white,
                    ]
                ]
                [
                    Personalized FL (\S \ref{sec:PersonalizedFL})
                    [
                        Category 1 (\S \ref{cat1}) \\
                        [
                            FedRep~\cite{collins2021exploiting}{, }
                            FED-ROD~\cite{chen2021bridging}{, }
                            FedBABU~\cite{oh2021fedbabu}
                            , leaf, text width=20em,
                            fill=cyan!20!white,
                        ]
                    ]
                    [
                        Category 2 (\S \ref{cat2})
                        [
                            Ditto~\cite{li2021ditto}{, }
                            FedPer~\cite{arivazhagan2019federated}
                            , leaf, text width=10em,
                            fill=purple!20!white,
                        ]
                    ]
                    [
                        Category 3 (\S \ref{cat3})
                        [
                            FedFomo~\cite{zhang2020personalized}{, }
                            FedBN~\cite{li2021fedbn}{, }
                            FedProto~\cite{tan2022fedproto}{, }
                            Apple~\cite{luo2022adapt}{, }
                            FedALA~\cite{zhang2023fedala}
                            , leaf, text width=30em,
                            fill=green!20!white,
                        ]
                    ]
                ]
            ]
        \end{forest}
    }
    \caption{Categorization of PFL algorithms based on learning and aggregation methods. 
    \parbox[t]{.45\linewidth}{\colorbox{cyan!10!white}{\textbf{Category 1}: Methods that learn a single global model and finetune it. }}  
    \parbox[t]{.45\linewidth}{\colorbox{purple!10!white}{\textbf{Category 2}: Methods that learn additional personalized models.}}
    \parbox[t]{.45\linewidth}{\colorbox{green!10!white}{\textbf{Category 3}: Methods that learn local models with personalized (local) aggregation.}}}
    \label{fig:sftframework}
\end{figure*}
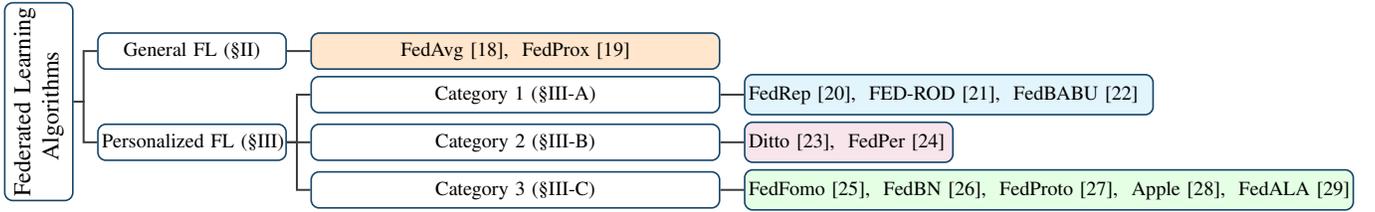

In this section,
we additionally categorize these existing pFL techniques \cite{collins2021exploiting, chen2021bridging, oh2021fedbabu, li2021ditto, arivazhagan2019federated, zhang2020personalized, li2021fedbn, tan2022fedproto, luo2022adapt, zhang2023fedala} into three categories as shown in Figure~\ref{fig:sftframework} based on learning and aggregation methodologies: (1) methods that learn a single global model and finetune it; (2) methods that learn additional personalized models; and (3) methods that learn local models with personalized (local) aggregation.

\subsection{Methods that learn a single global model and finetune it}
\label{cat1}
This approach focuses on methods that learn a single global model and then perform fine-tuning locally. Per-FedAvg utilizes an initial shared model based on MAML ~\cite{finn2017model}, while FedRep ~\cite{collins2021exploiting} splits the backbone into a global representation and a client-specific head, enabling personalized finetuning.

\subsubsection{FedRep}
\label{FedRep}
FedRep \cite{collins2021exploiting} learns a shared global representation using the Method of Moments. Each client then fine-tunes its local head with its own data. The local heads are aggregated to update the global representation, and this process is repeated over multiple epochs. By using a shared global representation and allowing for individual finetuning, FedRep balances capturing common features and accommodating personalized models.

Initially, FedRep learns a representation through the Method of Moments. In each communication round, a subset of clients updates their local head and the global representation by sampling fresh batches from their local distributions. The server updates the global representation by averaging the updated heads from clients, weighted by the number of samples used in each update. This client-server update process is repeated for multiple epochs.
The loss function in FedRep is a weighted sum of the losses over a representative subset of samples:
\begin{equation}
L_i(w) = \sum_{x_j \in D_i} w_{i,j} l(w, x_j, y_j)
\end{equation}

where $D_i$ is the representative subset from client $i$, $w_{i,j}$ is the weight of sample $j$ from client $i$, $l$ is the loss function, $w$ are the model parameters, $x_j$ is the input sample, and $y_j$ is the label.
The weight for each sample is:

\begin{equation}
w_{i,j} = \frac{p(x_j)}{K \cdot n_i}
\end{equation}

where $p(x_j)$ is the probability of sample $j$, $K$ is a normalization constant, and $n_i$ is the size of the subset for client $i$. The global model update is:

\begin{equation}
w_{t+1} = \frac{\sum_{i=1}^m n_i w_{t, i} w_{t, i}^{loc}}{\sum_{i=1}^m n_i w_{t,i}}
\end{equation}

where $w_{t,i}^{loc}$ is client $i$'s local model update at iteration $t$.

FedRep allows for more local updates per client by reducing the problem dimension with a shared representation. This leads to improved individual model accuracy and robustness to statistical heterogeneity among clients. However, FedRep may not be ideal for highly specialized tasks requiring unique features, potentially resulting in suboptimal performance.

\subsubsection{FED-ROD}
\label{fedrod}
Federated Robust Decoupling (FED-ROD) \cite{chen2021bridging} bridges the gap between global and personalized models for image classification tasks. It decouples the prediction task into two parts: a generic predictor trained on a large, diverse dataset shared by all clients, and a personalized predictor trained on each client's local data. FED-ROD achieves this by introducing a family of losses that are robust to non-identical class distributions, enabling clients to train a generic predictor with a consistent objective. Additionally, FED-ROD formulates the personalized predictor as an adaptive module that minimizes each client's empirical risk on top of the generic predictor.

One of the key advantages of FED-ROD is its ability to simultaneously achieve state-of-the-art performance in both generic and personalized FL tasks. By training a generic predictor with a consistent objective, FED-ROD improves the accuracy of the generic predictor and reduces the amount of data required for personalized training. The framework also offers flexibility and efficiency, as it can be integrated into existing FL systems and is compatible with a wide range of algorithms. The framework also includes a balanced risk minimization algorithm that encourages the local models to classify all classes well, which can improve the accuracy of the generic model and reduce the amount of data needed for personalized training.

Also, FED-ROD's adaptive module, which can be implemented as a linear or hypernetwork, allows for fast adaptation to new clients, further enhancing the performance of personalized models. However, the hypernetwork implementation of the adaptive module may require additional training and computation compared to the linear implementation.

\subsubsection{FedBABU}
\label{fedbabu}
FedBABU \cite{oh2021fedbabu} decouples the model parameters into the body and head parameters and only updates the body parameters during FL. The fixed random head can be shared across all clients, and the body is personalized to each client during the evaluation process by fine-tuning the head. Therefore, FedBABU learns local models with personalized aggregation by fine-tuning the head on each client's data to achieve personalization. During local updates, only the body parameters are trained using a randomly initialized head, while the head remains fixed throughout the federated training process. This design eliminates the need to aggregate the head and allows for more efficient communication and computation. The fixed random head can deliver comparable performance to a learned head in centralized settings. FedBABU further enables rapid personalization of the global model by fine-tuning the body of each client's data using the fixed random head.

One of the key advantages of FedBABU is its ability to reduce communication overhead and improve training efficiency. By focusing updates solely on the body parameters and sharing a fixed random head, the algorithm minimizes the information exchanged between clients and the server. This reduction in communication leads to faster training and lower resource requirements. On the other side, as the head remains fixed throughout training, there is limited room for personalized optimization of the head on a per-client basis. Some clients may benefit from a customized head, but FedBABU's fixed random head approach restricts personalization.

\vspace{-1em}
\begin{tcolorbox}
[breakable,width=0.48\textwidth,title={Takeaway:},boxrule=.3mm,colback=white,coltitle=white,left=.5mm, right=.5mm, top=.5mm, bottom=.5mm]    
   \emph{Finetuning methods start with a shared global model, then personalize it through local finetuning. This balances collaborative learning with individual adaptation, ideal for clients with similar base knowledge but needing customization.}
\end{tcolorbox}  

\begin{table*}[htbp]
\footnotesize
\centering
\caption{Analysis results of evaluation metrics and settings in Personalized Federated Learning (PFL) research.}
\label{tab1}
\rowcolors{2}{gray!20}{white}
\resizebox{\textwidth}{!}{
\begin{tabular}{lllllll}
\hline
\rowcolor{purple!10}
\textbf{Methods} & \textbf{Accuracy} & \textbf{Convergence} & \textbf{Comm. Cost} & \textbf{Time Cost} & \textbf{Datasets}$^\ast$ & \textbf{Comparison}$^\ddag$\\
\hline
\rowcolor{yellow!20}
\multicolumn{7}{c}{\textbf{General Federated Learning}}\\
\hline
\textbf{FedAvg} & Yes & No & Yes & No & (1), (7) &  \\
\textbf{FedProx} & Yes & Yes & No & No & (1), (4), (6), (7) & (a)\\
\hline
\rowcolor{yellow!20}
\multicolumn{7}{c}{\textbf{Personalized Federated Learning}}\\
\hline
\textbf{FedPer} & Yes & No & No & No & (2), (3), (8) & (a)\\
\textbf{FedFomo} & Yes & No & No & No & (1), (2), (3) & (a), (b), (c), (d), (e), (f), (g)\\
\textbf{Ditto} & Yes & Yes & No & No & (4), (5) & (f), (g), (h), (i), (j), (k), (l)\\
\textbf{FedBN} & Yes & Yes & No & No & (1), (9), (10), (11), (12) &  (a), (b)\\
\textbf{FedRep} & Yes & Yes & No & No & (2), (3), (4), (6) & (a), (b), (c), (f), (h), (i)\\
\textbf{FedBABU} & Yes & No & No & No & (3) & (a), (c), (f), (m), (n), (o)\\
\textbf{FED-ROD} & Yes & No & No & No & (13), (14) & (a), (b), (c), (f), (g), (m), (n), (o), (q), (r), (s)\\
\textbf{APPLE} & Yes & Yes & No & No & (1), (2) & (a), (b), (t), (u)\\
\textbf{FedProto} & Yes & Yes & Yes & No & (1), (2), (4) & (a), (b), (n), (o), (y)\\
\textbf{FedALA} & Yes & No & Yes & No & (1), (2), (3), (15), (16) & (a), (b), (f), (g), (m), (n), (t), (u), (v), (w), (x)\\
\hline
\end{tabular}
}
\caption*{\small \textbf{Datasets$^\ast$:} (1) MNIST, (2) CIFAR-10, (3) CIFAR-100, (4) FEMNIST, (5) CelebA, (6) Sent140, (7) Shakespeare, (8) FLICKR-AES, (9) SVHN, (10) USPS Hull, (11) SynthDigits, (12) MNIST-M, (13) FMNIST, (14) EMNIST, (15) Tiny-ImageNet, (16) AG News. \\
\textbf{Comparison$^\ddag$:} (a) FedAvg, (b) FedProx, (c) LG-FedAvg, (d) MOCHA, (e) CFL, (f) Per-FedAvg, (g) pFedMe, (h) APFL, (i) L2SGD, (j) Symmetrized KL, (k) EWC, (l) MAPPER, (m) Ditto, (n) FedRep, (o) FedPer, (q) SCAFFOLD, (r) FedDYN, (s) FedMTL, (t) FedFomo, (u) FedAMP, (v) FedPHP, (w) PartialFed, (x) APPLE, (y) FedSem.}
\end{table*}

\subsection{Methods that learn additional personalized models}
\label{cat2}
Another avenue explores methods that learn additional personalized models. pFedMe ~\cite{t2020personalized} adopts Moreau envelopes to learn personalized models for each client, whereas Ditto lets clients learn their personalized models through a proximal term that fetches information from the global model.

\subsubsection{Ditto}
\label{ditto}
In Ditto \cite{li2021ditto}, each device trains a personalized model using its local data and the global model as initialization. The resulting personalized models are used alongside the global model for predictions. These personalized models are regularized towards the optimal global model, balancing the trade-off between global and personalized objectives.

Ditto optimizes a global model and personalized models for each device in a statistically heterogeneous network. The optimization problem aims to minimize the global objective function, $G(w)$, which represents the regularized empirical risk minimization problem, and the personalized objective function, $H_k(v_k; w)$, which captures the local data distribution of each device. The trade-off between global and personalized objectives is controlled by the hyperparameter $\lambda$. The optimization process uses an alternating algorithm where the global model is updated by aggregating the personalized models from all devices, and the personalized models are updated by minimizing the local objective function on each device. Ditto offers several advantages, including capturing statistical heterogeneity, balancing the trade-off between global and personalized objectives, and providing fairness, robustness, scalability, and flexibility in model selection.

\begin{equation}
\underset{v_{k}}{\text{min}} = h_k(v_k; w^*):=F_k (v_k) + \frac{\lambda}{2}||v_k - w^*||^2
\end{equation}
\begin{equation}
\text{s.t.} \quad w^* \in \text{arg}\min_w G(F_1(k), \ldots, F_K(w))
\end{equation}

Ditto begins by initializing the global model and personalized models. The global model is then communicated to all devices, which train their personalized models using their local data and the global model as initialization. The personalized models are sent back to the server for aggregation. This communication and aggregation process is repeated for multiple rounds until convergence or a stopping criterion is met. Throughout the iterations, the alternating optimization algorithm updates the global model and personalized models, balancing objectives.

One advantage of Ditto is its ability to capture the statistical heterogeneity of each device's data distribution. The personalized models offer better accuracy and generalization compared to homogeneous approaches. Another advantage is the regularization term, which encourages the personalized models to be close to the optimal global model. This regularization strikes a balance between global and personalized objectives, ensuring consistency while accounting for local variations. Ditto also provides fairness and robustness by mitigating tensions between these constraints. However, Ditto has some drawbacks, including the communication and computation resources required to train and update the personalized models, which may increase the communication overhead and computational cost. Additionally, the optimal value of the hyperparameter $\lambda$ may be sensitive to the specific dataset and task.

\subsubsection{FedPer}
\label{fedper}
FedPer \cite{arivazhagan2019federated} addresses the challenge of statistical heterogeneity in non-identical data partitions by training a shared base model on the server and personalizing it for each client using client-specific parameters. This personalization is achieved through the addition of personalized layers to the base model, allowing it to capture client-specific features.

The algorithm consists of two phases: global aggregation phase and local update phase. In the global aggregation phase, the server aggregates the model updates from the clients using a federated averaging approach. In the local update phase, each client updates its base and personalized layers locally using a stochastic gradient descent (SGD) style algorithm. The personalization layers capture the client-specific features, which can improve the accuracy of the model for each client. The FedPer algorithm uses a loss function that combines global loss and local loss for each client. The global loss ensures that the global model is aligned with the overall dataset, while the local loss ensures that the local models capture the nuances of each client's data. Mathematically, the FedPer loss function is:

\begin{equation}
L(\theta) = \sum_{k=1}^{K} w_k L_k(\theta) + \lambda R(\theta)
\end{equation}

where $\theta$ represents the model parameters, $K$ is the number of clients, $w_k$ is the weight assigned to client $k$, $L_k(\theta)$ is the local loss for client $k$, $R(\theta)$ is a regularization term, and $\lambda$ is the regularization parameter.

By incorporating personalized layers, FedPer can capture client-specific characteristics, leading to more accurate models for individual clients. The approach is also versatile and can be applied to various deep-learning model families and datasets, as demonstrated by the authors. However, FedPer has some limitations. It requires a large number of clients to be effective since the personalization layers are trained on a per-client basis. 

\begin{tcolorbox}[breakable,width=0.48\textwidth,title={Takeaway:},boxrule=.3mm,colback=white,coltitle=white,left=.5mm, right=.5mm, top=.5mm, bottom=.5mm]    
   \emph{Methods that learn personalized algorithms maintain both global and client-specific models. This dual-model approach offers flexibility in capturing client nuances while leveraging collective knowledge, making it well-suited for applications with significant data heterogeneity across clients.}
\end{tcolorbox}

\subsection{Methods that learn local models with personalized (local) aggregation.}
\label{cat3}
Recent research also investigates methods that aim to capture personalization by generating client-specific models through personalized aggregation, contributing to better local models with personalized elements.

\subsubsection{FedFomo} 
\label{fedfomo}
Instead of learning a single global model or generating additional personalized models, FedFomo \cite{zhang2020personalized} learns local models on each client's data. The personalized aggregation step allows for the computation of personalized weighted combinations of the models, resulting in stronger models tailored to the individual client's objectives. By prioritizing personalization and leveraging locally trained models, FedFomo distinguishes itself within the field of FL.

FedFomo aims to optimize models for individual client objectives. Unlike traditional FL methods, FedFomo does not compute a single global model. Instead, it introduces two new steps: (1) ranking the models based on their suitability towards the client's objective, and (2) computing personalized weighted combinations of the downloaded models. This approach allows each client to obtain a stronger personalized model.

FedFomo learns optimal combinations of the available server models for each participating client. For this client, information is leveraged in two ways. First, the aim is to optimize directly for each client's target objective. It assumes that clients can distinguish between good and bad models on their target tasks through a labeled validation data split $D_{i}^{val}$ $\subset$ $D_i$ into the client's local data. The client can then evaluate any arbitrary model $\theta_j$ on this validation set and quantify the performance through the computed loss, denoted by $L_i(\theta_j)$. 

Second, they directly leverage the potential heterogeneity among client models, similar to ~\cite{zhao2018federated}, where they show that diverging model weights come directly from local data heterogeneity. However, instead of combining these parameters into a single global model, uploaded models are maintained individually to preserve a model's potential contribution to another client. These ideas enable clients to optimize data distributions different from their own local data.

One of the key advantages of FedFomo is its emphasis on personalization. By allowing clients to optimize their models for their own specific objectives, FedFomo produces stronger and more effective models that are tailored to individual needs. However, the introduction of two additional steps in the traditional FL process increases the complexity of the framework. This complexity may require additional resources and expertise to implement and manage effectively. 


\subsubsection{FedBN}
\label{fedbn}
FedBN \cite{li2021fedbn} trains individual local models on each client using their own data and local BN parameters. The personalized (local) aggregation scheme at the server preserves the local data distribution while combining the models. FedBN achieves this by maintaining independent BN parameters, which effectively mitigate feature shifts and enhance the convergence behavior and performance on non-IID datasets.

The main difference between FedBN and FedAvg is that FedBN uses local batch normalization (BN) to mitigate feature shifts in non-IID data. However, in traditional FL, the local BN parameters are synchronized with the global model, which can lead to feature shifts in non-IID data. In FedBN, the local BN parameters are not synchronized with the global model, which helps to preserve the local data distribution and mitigate feature shifts. Another difference is that FedBN is independent of the communication and aggregation strategy and can be combined with different optimization algorithms, communication schemes, and aggregation techniques. In contrast, FedAvg is based on a specific communication and aggregation strategy, where the server aggregates client updates using simple averaging.

The advantage of FedBN is its compatibility with existing FL toolkits and systems. Being a lightweight modification to FedAvg, it can be easily integrated into various optimization algorithms, communication schemes, and aggregation techniques. However, FedBN does have certain limitations. It assumes that local models have BN layers, which may not always be the case in real-world scenarios. This restricts its applicability to models that utilize BN layers.

\subsubsection{FedProto}
\label{fedproto}
FedProto \cite{tan2022fedproto} allows clients to train their own local models, which are then aggregated using abstract class prototypes exchanged between the server and clients. By leveraging local aggregation, FedProto balances the benefits of personalized models with the need for global coordination.

FedProto employs abstract class prototypes instead of gradients to enhance optimization convergence and generalization performance. It begins with the server initializing global prototypes for each class. Each client then trains its local model on its own data, minimizing classification error while maintaining proximity to the global prototypes. Local prototypes are sent by each client to the server, which aggregates them to obtain the global prototypes. Finally, the server broadcasts the global prototypes to all clients for the next round, and these steps are repeated until convergence or a stopping criterion is met.

One of the main advantages of FedProto is its improved communication efficiency. By using abstract class prototypes instead of gradients, the communication overhead is reduced, making the framework more efficient. FedProto also prioritizes privacy preservation since the prototypes are generated by averaging low-dimensional representations of samples from the same class. This protects data privacy while allowing for effective communication. Additionally, FedProto handles heterogeneity across clients by regularizing local models with global prototypes, leading to improved generalization performance. However, FedProto does have certain limitations. The generation of abstract class prototypes requires additional computation and storage, which may increase computational and storage overhead. Furthermore, the quality of the abstract class prototypes depends on the quality of the low-dimensional representations, which can be influenced by the feature extractor.

\subsubsection{APPLE}
\label{apple}
APPLE \cite{luo2022adapt} addresses non-IID data challenges through adaptive learning of directed relationships (DRs) and the construction of core models. The core idea behind APPLE is to adaptively learn the quantification of how much each client can benefit from other clients' models, referred to as directed relationships (DRs). 

In APPLE, after local training, instead of uploading the entire personalized model, each client uploads a constructing component called a core model to the central server. The server maintains these core models for further downloading purposes. To achieve personalized models that take advantage of beneficial core models while suppressing less beneficial ones, APPLE adaptively learns a unique set of local weights on each client known as DR vectors. These DR vectors are used to weigh the downloaded core models for flexible control over the focus of training between global and local objectives.

Advantages of APPLE include by adaptively learning DR vectors and weighting core models, APPLE achieves better personalization and mitigates the distribution shift across non-IID datasets. Additionally, APPLE reduces privacy risks by uploading only the core models instead of the personalized models, thus protecting sensitive client data. However, the approach lead to increased communication overhead due to the learning of DR vectors on each client, potentially resulting in slower convergence and increased training time. The increased complexity of the APPLE approach pose implementation and maintenance challenges compared to other pFL methods.

\subsubsection{FedALA}
\label{fedala}
FedALA's \cite{zhang2023fedala} Adaptive Local Aggregation (ALA) module exemplifies personalized (local) aggregation by adaptively aggregating the global and local models towards the local objective of each client, tailoring the global model to individual client characteristics. This approach improves model customization and enhances the performance of FL in the presence of statistical heterogeneity.

FedALA addresses the statistical heterogeneity challenge by utilizing ALA module. The algorithm begins with the central server initializing the global model parameters and broadcasting them to all clients. Each client then initializes its local model by adaptively aggregating the downloaded global model and local model towards the local objective using the ALA module. The clients train their local models on their private data for a fixed number of epochs. The updated local models are then aggregated at the central server, and this process is repeated for multiple iterations or until convergence is achieved.

The key advantage of FedALA is its ability to personalize the global model by adaptively aggregating the global and local models towards the local objective of each client. This allows it to capture the desired information at the element level, improving the performance of existing FL methods. By addressing statistical heterogeneity, it can potentially achieve higher accuracy and convergence rates compared to traditional FL algorithms. However, the ALA module introduces complexity and computational overhead due to fine-grained element-wise aggregation and real-value weight learning. Lastly, FedALA, like many ML methods, is sensitive to hyperparameters, requiring careful tuning for optimal performance, which can be time-consuming and require expertise.

\begin{tcolorbox}[breakable,width=0.48\textwidth,title={Takeaway:},boxrule=.3mm,colback=white,coltitle=white,left=.5mm, right=.5mm, top=.5mm, bottom=.5mm]    
   \emph{pFL techniques based on personalized (local) aggregation use personalized aggregation to combine knowledge, emphasizing local data characteristics while enabling some knowledge sharing. This is useful when clients have specialized data that doesn't benefit from global averaging.}
\end{tcolorbox}


\section{Evaluation}
\label{sec:Results}

\subsection{Evaluation Questions}
\label{sec:Questions}
We show the metrics evaluated in the original papers of these pFL works. All of the original papers on these pFL methods use different metrics and datasets as shown in Table~\ref{tab1}. This highlights a lack of a standard benchmark for comparative evaluation of these techniques. Therefore, we use a standard benchmark with same datasets and metrics in controlled settings to answer the following research questions:
\\
\textbf{Q1:} How do the different categories of personalized FL methods compare in terms of performance and speed?
\\
\textbf{Q2:} Which algorithms are most and least memory-intensive?\\
\textbf{Q3:} Which algorithm stands out as top performer, and why?\\
\textbf{Q4:} What are the accuracy-memory overhead tradeoffs among personalized FL algorithms?

\subsection{Evaluation Metrics}

\begin{itemize}
\item \textbf{Accuracy ($Acc$)}: 
Accuracy measures the proportion of correctly predicted labels by the model at convergence. It indicates how well the model adapts to each client's data while maintaining global performance.

\item \textbf{AUC ($AUC$)}: 
The Area Under the Curve (AUC) at convergence evaluates the model's performance based on the ROC curve, which illustrates the trade-off between a true positive rate and a false positive rate. This metric is significant in personalized FL because it allows us to assess the model's ability to handle imbalanced data which is the case for non-IID distribution.

\item \textbf{Communications Rounds Required for Convergence ($R_{Conv}$)}: 
The number of communication rounds required for convergence indicates the efficiency of the personalized FL algorithm. Reducing this number is important to minimize communication overhead and enhance overall FL efficiency.

\item \textbf{Time Taken for Convergence ($T_{Conv}$)}:
The time taken for the personalized FL model to converge is a metric that highlights the speed at which the model adapts to local client data while achieving a globally acceptable performance level. Faster convergence is an advantage in personalized FL.

\item \textbf{Time Taken to Complete 200 Rounds($T_{200}$)}:
The time taken to complete the 200 communication rounds. This offers insight into the overall computational efficiency of the personalized FL system. 

\item \textbf{Memory}:
Memory usage indicates the memory requirements of the algorithm during execution. Personalized FL systems deal with multiple clients, each with their data and local models, which can lead to significant memory consumption. Understanding memory usage helps in resource management and scalability of the FL system.

\end{itemize}


\subsection{Experimental Settings}
We use MNIST and FMNIST datasets which have 28$\times$28 size. To simulate a realistic scenario in a distributed environment, we split the data among 20 users using the Dirichlet distribution. The joining ratio was set to 5, meaning only 5 out of the 20 users participated in each round of FL. The learning rate was set to 0.05, and the threshold time for clients to communicate with the aggregator was set to 500 seconds. We adopted the hyperparameters from the original papers to ensure consistency and comparability of our results.

\subsection{Experimental Results}
The results for two baselines and ten pFL techniques for MNIST (DS-1), MNIST (DS-2), FMNIST (DS-1), and FMNIST (DS-2) are presented in Table~\ref{tab2}, Table~\ref{tab3}, Table~\ref{tab4}, and Table~\ref{tab5} respectively. The best result for each metric is highlighted in bold.

\begin{figure}[htbp]
\centering
\begin{subfigure}
    \centering
    \includegraphics[width=0.8\linewidth]{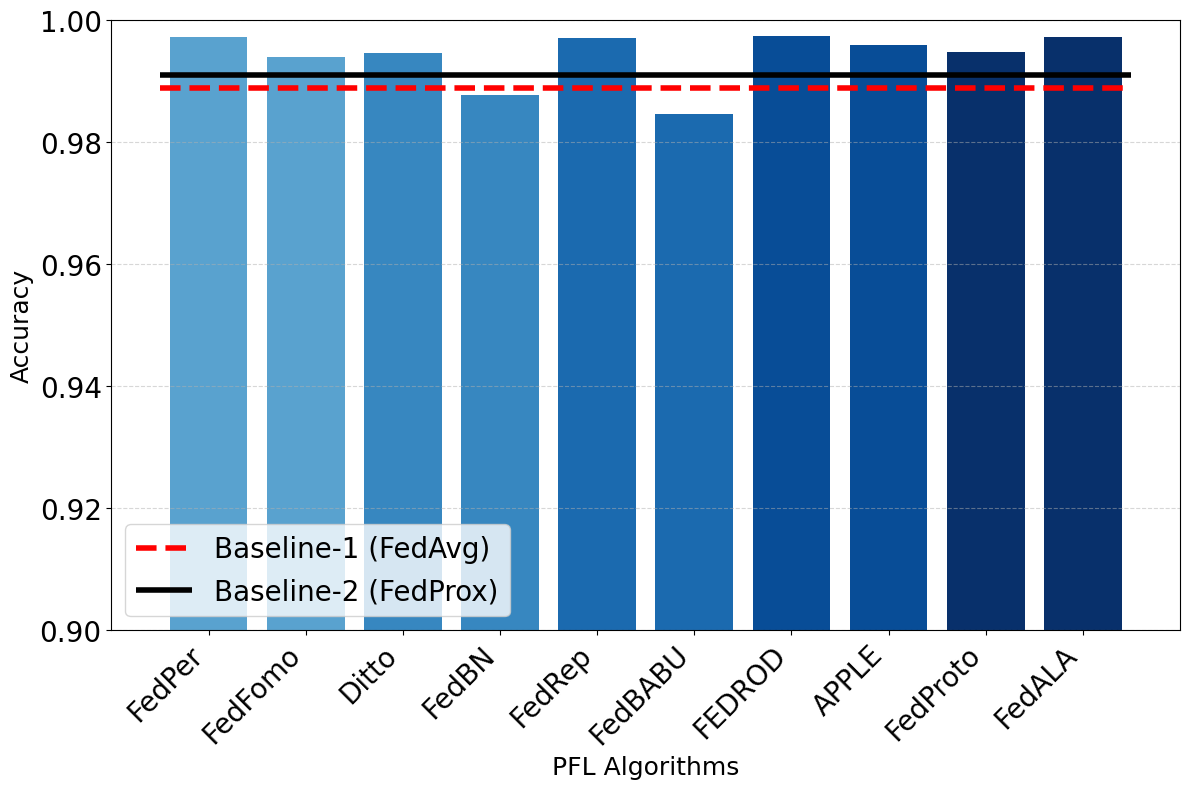}
    \caption{Accuracy on MNIST DS-1.}
    \label{AccMNIST1}
\end{subfigure}
\vfill
\begin{subfigure}
    \centering
    \includegraphics[width=0.8\linewidth]{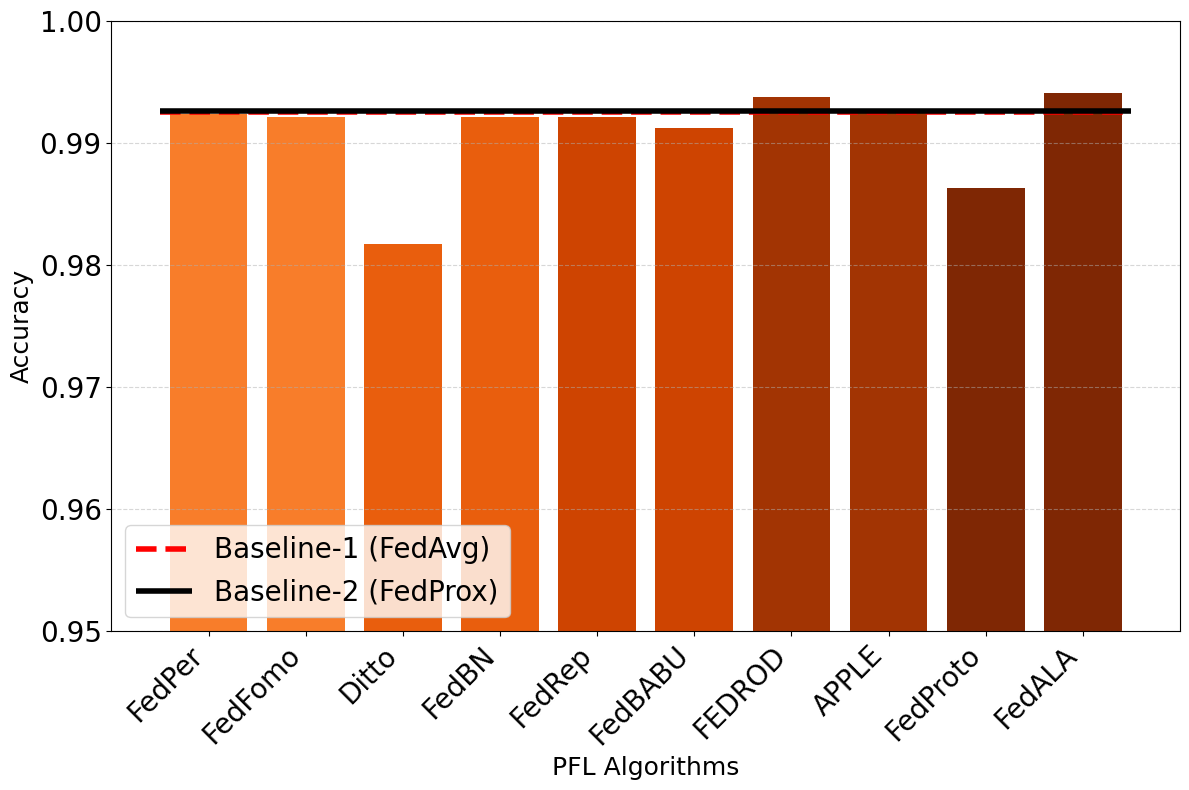}
    \caption{Accuracy on MNIST DS-2.}
    \label{AccMNIST2}
\end{subfigure}
\vfill
\begin{subfigure}
    \centering
    \includegraphics[width=0.8\linewidth]{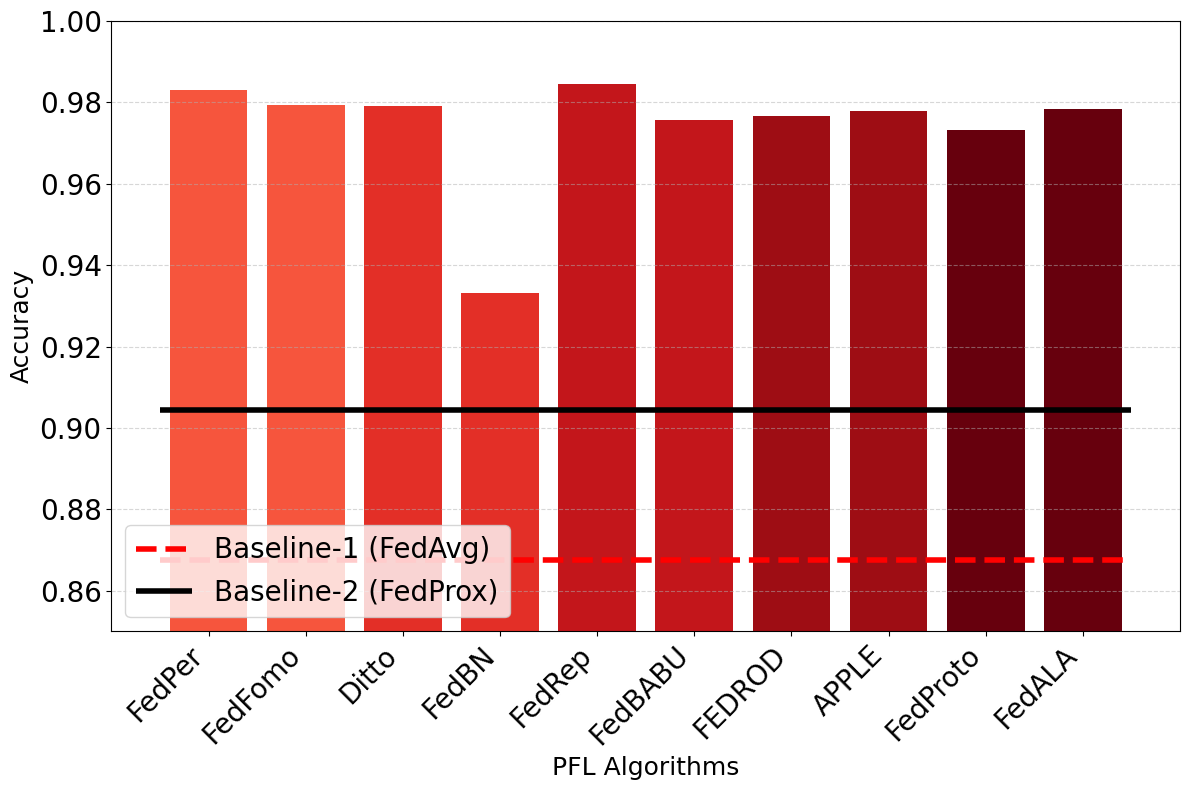}
    \caption{Accuracy on FMNIST DS-1.}
    \label{AccFMNIST1}
\end{subfigure}
\vfill
\begin{subfigure}
    \centering
    \includegraphics[width=0.8\linewidth]{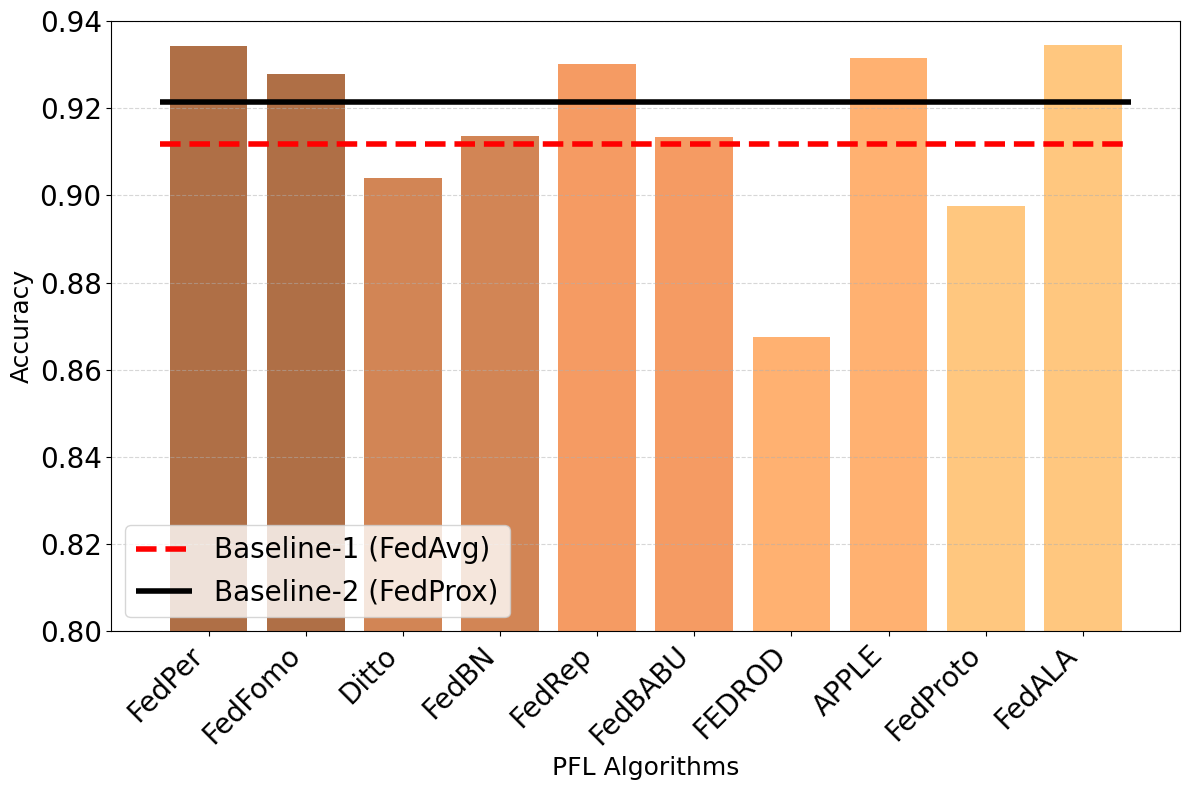}
    \caption{Accuracy on FMNIST DS-2.}
    \label{AccFMNIST2}
\end{subfigure}
\caption{Accuracies of personalized-FL techniques for MNIST and FMNIST datasets.}
\label{Acc}
\end{figure}

Figures~\ref{AccMNIST1},~\ref{AccMNIST2}, \ref{AccFMNIST1}, and \ref{AccFMNIST2} depict the accuracy comparison between pFL algorithms and the two baseline methods for MNIST (DS-1), MNIST (DS-2), FMNIST (DS-1), and FMNIST (DS-2) datasets. The results demonstrate that FedPer, FedRep, FEDROD, and FedALA consistently outperformed the two baselines across all datasets. Notably, all pFL algorithms exhibited superior performance even in the most challenging data scenario, FMNIST(DS-1), as illustrated in Table~\ref{tab4} and Figure~\ref{AccFMNIST1}.

\begin{figure}[htbp]
\centering
\begin{subfigure}
    \centering
    \includegraphics[width=0.8\linewidth]{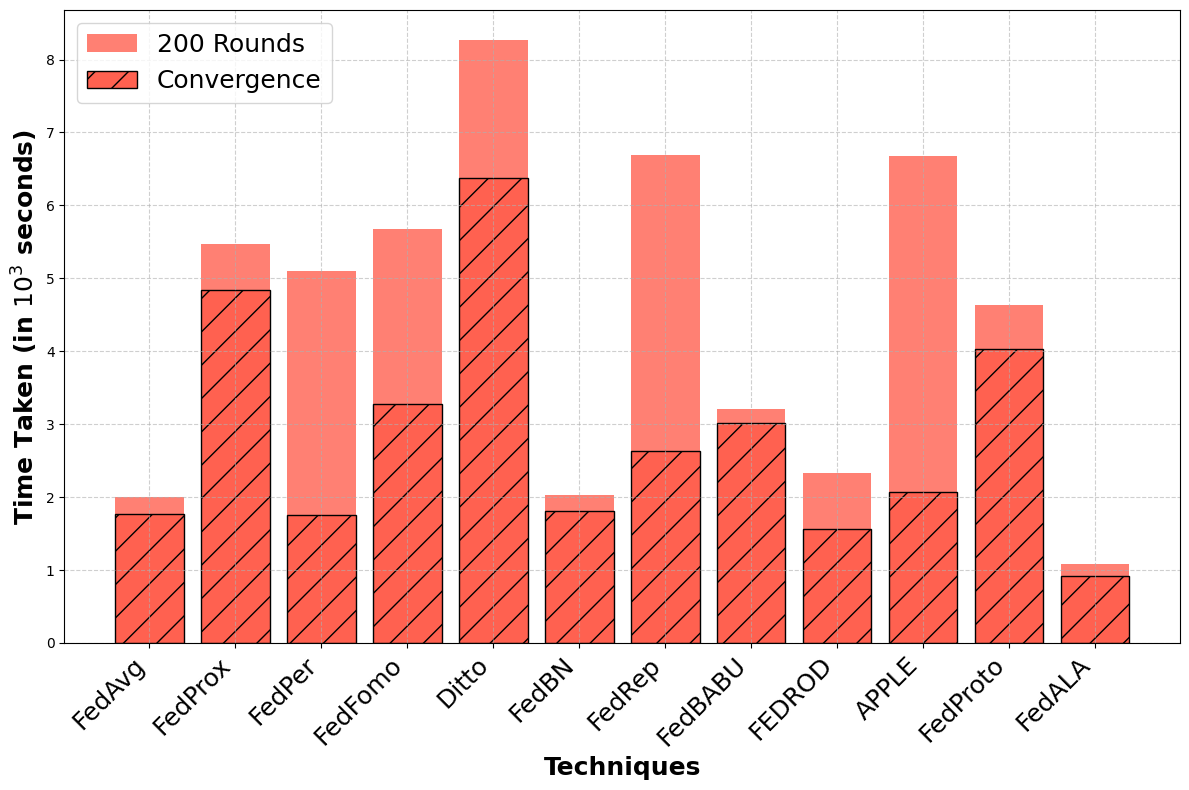}
    \caption{Comparison of convergence and 200 rounds time across techniques.}
    \label{time}
\end{subfigure}
\vfill
\begin{subfigure}
    \centering
    \includegraphics[width=0.8\linewidth]{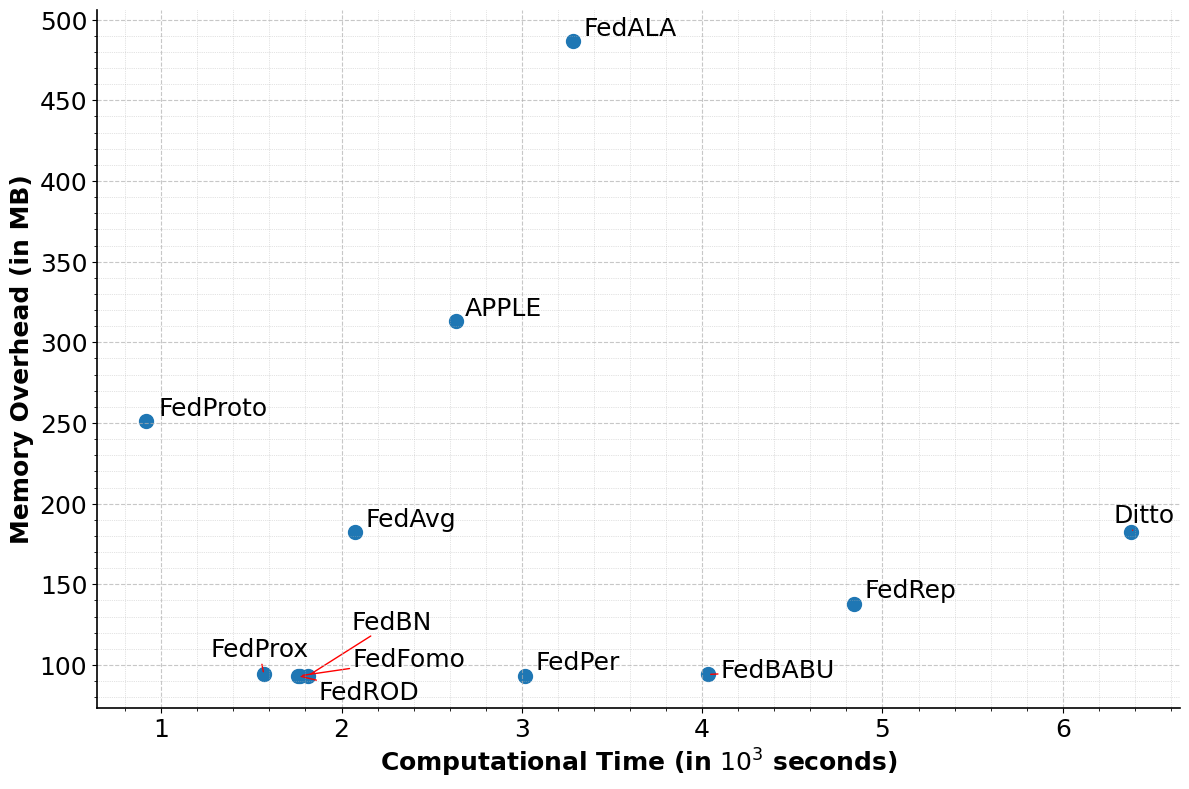}
    \caption{Time taken vs Memory Overhead}
    \label{timevscomm}
\end{subfigure}
\vfill
\begin{subfigure}
    \centering
    \includegraphics[width=0.8\linewidth]{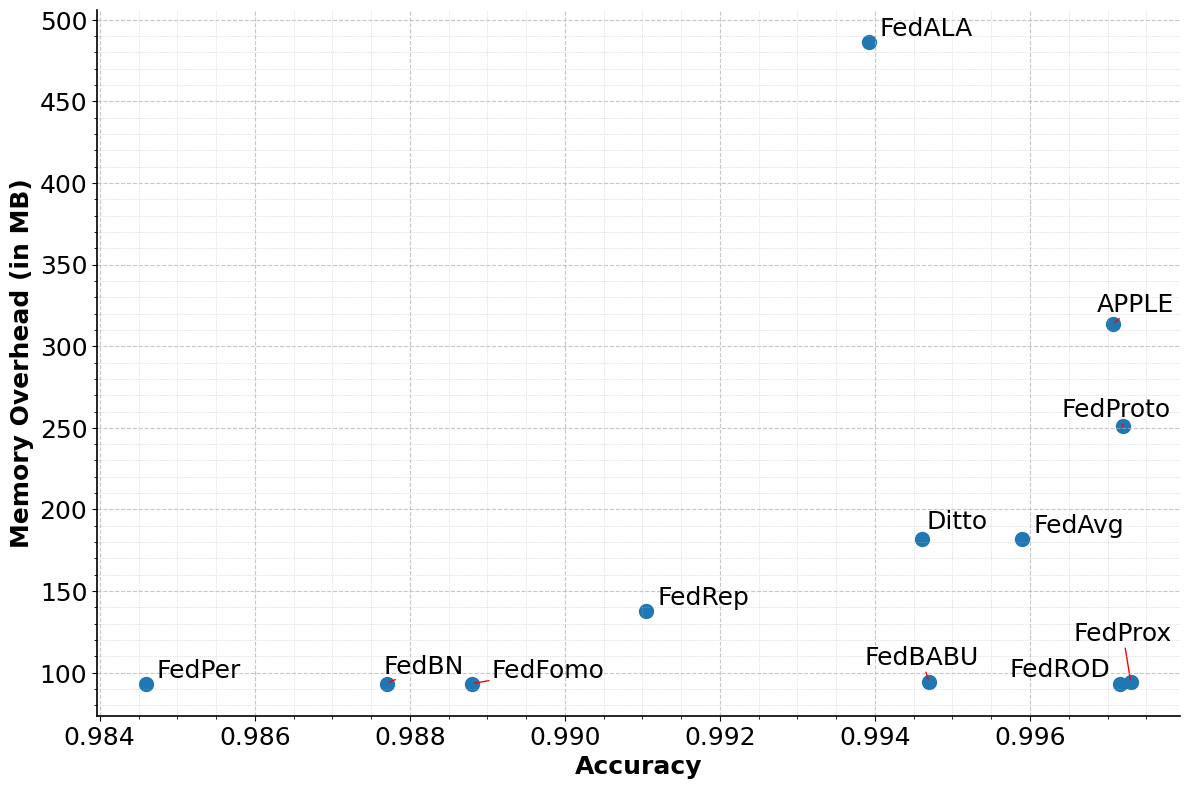}
    \caption{Accuracy vs Memory Overhead}
    \label{accComm}
\end{subfigure}
\caption{Performance of different pFL techniques}
\label{Exps}
\end{figure}

The graph in Figure~\ref{time} illustrates the performance of baselines and pFL algorithms concerning convergence and completion over 200 rounds. The data represents the average results across all four settings. Analyzing the time taken for convergence provides insights into how swiftly an algorithm reaches convergence while considering the time for 200 rounds reveals the time per round. Notably, FedALA outperforms all other algorithms in both metrics. FedBN and FEDROD also exhibit competitive performance, being closely aligned with baseline-1. However, Ditto stands out as the most time-consuming approach for both convergence and completing 200 rounds. The relatively slower convergence of Ditto compared to other methods can be attributed to its requirement of learning additional personalized models and balancing the trade-off between global and personalized objectives. Consequently, the time taken per round is extended due to the need for training these additional personalized models.
\begin{table}[htbp]
\footnotesize
\centering
\caption{MNIST DS-1 Performance Metrics}
\label{tab2}
\rowcolors{2}{gray!20}{white}
\begin{tabular}{lcccccc}
\hline
\rowcolor{purple!10}
\textbf{Technique} & \textbf{Acc} & \textbf{AUC} & \textbf{$R_{Conv}$} & \textbf{$T_{Conv}$(s)} & \textbf{$T_{Total}$(s)} & \textbf{Mem.}(MB) \\
\hline
Baseline-1 & 0.99 & 0.999 & 196 & 1884 & 1923 & 93.31 \\
Baseline-2 & 0.99 & 0.999 & 187 & 5109 & 5491 & 137.74 \\
FedFomo & 0.99 & 0.999 & 56 & 1576 & 5782 & 486.56 \\
Ditto & 0.99 & 1.00 & 149 & 5529 & 7464 & 182.18 \\
FedRep & \textbf{1.00} & \textbf{1.00} & \textbf{55} & 1902 & 6951 & 313.46 \\
FedPer & 1.00 & 1.00 & 88 & 2281 & 5210 & \textbf{93.29} \\
FedBABU & 0.98 & 0.999 & 181 & 1861 & 2067 & \textbf{93.29} \\
FEDROD & \textbf{1.00} & \textbf{1.00} & 158 & 1893 & 2394 & 94.11 \\
FedALA & 1.00 & 0.999 & 154 & \textbf{825} & \textbf{1070} & 251.11 \\
APPLE & 1.00 & 0.999 & 34 & 1155 & 6638 & 182.18 \\
FedBN & 0.99 & 0.999 & 191 & 1914 & 2004 & 93.31 \\
FedProto & 0.99 & N/A & 156 & 3532 & 4522 & 94.12 \\
\hline
\end{tabular}
\end{table}

\begin{table}[htbp]
\footnotesize
\centering
\caption{MNIST DS-2 Performance Metrics}
\label{tab3}
\rowcolors{2}{gray!20}{white}
\begin{tabular}{lcccccc}
\hline
\rowcolor{purple!10}
\textbf{Technique} & \textbf{Acc} & \textbf{AUC} & \textbf{$R_{Conv}$} & \textbf{$T_{Conv}$(s)} & \textbf{$T_{Total}$(s)} & \textbf{Mem.}(MB) \\
\hline
Baseline-1 & 0.99 & 0.999 & 131 & 1285 & 1857 & 93.31 \\
Baseline-2 & 0.99 & 0.999 & 176 & 4660 & 5322 & 137.74 \\
FedFomo & 0.99 & 0.999 & 135 & 3689 & 5456 & 479.89 \\
Ditto & 0.98 & 1.00 & 186 & 7895 & 8606 & 182.18 \\
FedRep & 0.99 & 0.999 & 98 & 3128 & 6415 & 313.46 \\
FedPer & 0.99 & 0.999 & 87 & 2057 & 4753 & \textbf{93.29} \\
FedBABU & 0.99 & 0.999 & 191 & 6412 & 6748 & \textbf{93.29} \\
FEDROD & 0.99 & 0.999 & 119 & 1518 & 2544 & 94.11 \\
FedALA & \textbf{0.99} & \textbf{1.00} & 152 & \textbf{822} & \textbf{1080} & 251.11 \\
APPLE & 0.99 & 0.999 & \textbf{54} & 1812 & 6748 & 182.18 \\
FedBN & 0.99 & 0.999 & 137 & 1367 & 2006 & 93.31 \\
FedProto & 0.99 & N/A & 187 & 4352 & 4654 & 94.12 \\
\hline
\end{tabular}
\end{table}

\begin{table}[htbp]
\footnotesize
\centering
\caption{FMNIST DS-1 Performance Metrics}
\label{tab4}
\rowcolors{2}{gray!20}{white}
\begin{tabular}{lcccccc}
\hline
\rowcolor{purple!10}
\textbf{Technique} & \textbf{Acc} & \textbf{AUC} & \textbf{$R_{Conv}$} & \textbf{$T_{Conv}$(s)} & \textbf{$T_{Total}$(s)} & \textbf{Mem.}(MB) \\
\hline
Baseline-1 & 0.87 & 0.98 & 185 & 1970 & 2129 & 93.31 \\
Baseline-2 & 0.90 & 0.98 & 152 & 4184 & 5533 & 137.74 \\
FedFomo & 0.98 & 1.00 & 127 & 3633 & 5730 & 493.22 \\
Ditto & 0.98 & 1.00 & 158 & 6528 & 8372 & 182.18 \\
FedRep & \textbf{0.98} & 1.00 & 60 & 1942 & 6505 & 313.46 \\
FedPer & 0.98 & 1.00 & 36 & 927 & 5179 & \textbf{93.29} \\
FedBABU & 0.98 & 0.98 & 188 & 1912 & 2034 & \textbf{93.29} \\
FEDROD & 0.98 & \textbf{1.00} & 78 & \textbf{896} & 2281 & 94.11 \\
FedALA & 0.98 & 1.00 & 160 & 905 & \textbf{1130} & 251.11 \\
APPLE & 0.98 & 1.00 & \textbf{55} & 1845 & 6623 & 182.18 \\
FedBN & 0.93 & 0.98 & 189 & 1932 & 2027 & 93.31 \\
FedProto & 0.98 & N/A & 183 & 3649 & 4623 & 94.12 \\
\hline
\end{tabular}
\end{table}

\begin{table}[htbp]
\footnotesize
\centering
\caption{FMNIST DS-2 Performance Metrics}
\label{tab5}
\rowcolors{2}{gray!20}{white}
\begin{tabular}{lcccccc}
\hline
\rowcolor{purple!10}
\textbf{Technique} & \textbf{$Acc$} & \textbf{$AUC$} & \textbf{$R_{Conv}$} & \textbf{$T_{Conv}$(s)} & \textbf{$T_{Total}$(s)} & \textbf{Mem.}(MB) \\
\hline
Baseline-1 & 0.89 & 0.97 & 198 & 1365 & 1958 & 93.31 \\
Baseline-2 & 0.89 & 0.97 & 154 & 4982 & 5611 & 137.74 \\
FedFomo & 0.96 & 1.00 & 163 & 4915 & 6355 & 499.76 \\
Ditto & 0.97 & 0.99 & 177 & 6524 & 7620 & 182.18 \\
FedRep & 0.97 & \textbf{1.00} & 132 & 2328 & 6584 & 313.46 \\
FedPer & 0.97 & 1.00 & 75 & 1657 & 5083 & \textbf{93.29} \\
FedBABU & 0.96 & 0.99 & 189 & 2782 & 3033 & \textbf{93.29} \\
FEDROD & 0.96 & 1.00 & 179 & \textbf{827} & \textbf{1082} & 94.11 \\
FedALA & \textbf{0.97} & 0.99 & 178 & 1028 & 1262 & 251.11 \\
APPLE & 0.96 & 0.99 & \textbf{54} & 1914 & 6682 & 182.18 \\
FedBN & 0.96 & 0.99 & 135 & 1762 & 1925 & 93.31 \\
FedProto & 0.96 & N/A & 173 & 3678 & 4571 & 94.12 \\
\hline
\end{tabular}
\end{table}

Based on the empirical analysis, we address our research questions from Section (\S \ref{sec:Questions}).

\textbf{A1: }Our experimental results reveal a clear hierarchy in performance and speed among the different categories of personalized FL methods (as shown in Tables~\ref{tab2}, \ref{tab3}, \ref{tab4}, \ref{tab5}). Category 3 methods demonstrate the fastest performance, followed by Category 1 and then Category 2 (as shown in Figure~\ref{time}) . This advantage of Category 3 can be attributed to its ability to reduce communication and computation requirements per round. In Category 3, each device trains its own local model using its local data and only communicates with a central server for aggregation, which can be done using efficient methods such as averaging or adaptive local aggregation, leading to faster convergence. As shown in Figure~\ref{timevscomm}, Category 2 methods are slower due to the need to train and maintain multiple models per client, increasing computational and memory demands. This can be challenging for resource-constrained clients, requiring a trade-off between global and personalized objectives.

In terms of accuracy, Category 3 outperforms Category 1, with Category 2 showing the lowest performance. Category 3's superior accuracy is due to its ability to effectively capture the heterogeneity and diversity of data across devices. By allowing each device to train its own local model and then aggregating these models using personalized (local) aggregation, Category 3 methods better represent the unique characteristics of each device's data distribution, resulting in improved generalization on new data. In contrast, Category 1 methods may struggle to capture individual characteristics, exhibiting bias towards certain devices in heterogeneous data environments.

\begin{tcolorbox}[breakable,width=0.48\textwidth,title={Takeaway:},boxrule=.3mm,colback=white,coltitle=white,left=.5mm, right=.5mm, top=.5mm, bottom=.5mm]    
   \emph{Category 3 pFL methods are fastest by using local model training and aggregation to handle data heterogeneity. Category 2 methods are slower due to the higher computational demands of maintaining multiple models per client.}
\end{tcolorbox} 

\textbf{A2: }Among the various algorithms tested, FedALA emerges as a top performer. This superiority is due to its implementation of an Adaptive Local Aggregation (ALA) module, which adaptively aggregates the downloaded global model and local model towards each client's local objective, initializing the local model before training in each iteration. FedALA's personalized (local) aggregation enables precise information capture at the element level, a notable advantage over methods relying on binary and layer-wise weight learning. This approach allows FedALA to effectively balance global knowledge with the specific characteristics of each client's local data, resulting in improved overall performance. The adaptive nature of FedALA's aggregation strategy allows it to dynamically adjust to client's needs, contributing to its exceptional performance across various FL scenarios as shown in Figures~\ref{AccMNIST1}, \ref{AccMNIST2}, \ref{AccFMNIST1}, \ref{AccFMNIST2}.


\textbf{A3: }Our findings indicate that FedFomo is the most memory-intensive algorithm (Figure~\ref{timevscomm}, while FedPer and FedBABU are the least memory-intensive. FedFomo's high memory usage stems from calculating optimal weighted combinations of models for each client. It downloads individual models from other clients to form these personalized combinations, requiring considerable memory to manage and maintain multiple model combinations. Conversely, FedPer demonstrates communication efficiency by transmitting only client-specific personalization parameters rather than entire model parameters, as shown in Figure~\ref{timevscomm}. This significantly reduces the data exchanged during each communication round, resulting in lower memory usage and communication costs. Similarly, FedBABU achieves memory efficiency by updating only the model body during federated training, leaving the head unchanged. This reduces the parameters communicated for training, as the head remains constant. These memory-efficient approaches make FedPer and FedBABU suitable for resource-constrained environments.

\textbf{A4: }Our experiments reveal notable accuracy-memory overhead tradeoffs among pFL algorithms (Figure~\ref{accComm}). FedPer, Apple, FedBABU, Ditto, Proto, and FEDROD show exceptional performance in balancing accuracy and memory overhead. These algorithms are favorable in the accuracy-memory overhead tradeoff plot (Figure~\ref{accComm}), achieving a balance between model performance and memory overhead. This makes them valuable for scenarios with limited communication bandwidth, ensuring efficient data exchange between clients and the central server.

FedPer, in particular, achieves high accuracy while maintaining communication efficiency by transmitting only client-specific personalization parameters, significantly reducing data exchanged per communication round. This efficiency does not compromise accuracy, as FedPer effectively captures client-specific nuances. Similarly, FedBABU updates only the model body during training, contributing to communication efficiency while maintaining competitive accuracy. These algorithms show that high model performance can be achieved without excessive memory overhead, a crucial consideration in real-world FL applications with limited network bandwidth. Our study's performance patterns provide a strong foundation for exploring accuracy-memory overhead tradeoffs in FL, opening avenues for future research to optimize these tradeoffs.

\begin{tcolorbox}[breakable,width=0.48\textwidth,title={Takeaway:},boxrule=.3mm,colback=white,coltitle=white,left=.5mm, right=.5mm, top=.5mm, bottom=.5mm]    
   \emph{Category 1 methods excel in the accuracy-memory overhead tradeoff due to their finetuning approach on local data, which allows for higher accuracy.}
\end{tcolorbox} 

\section{pFL: New Opportunities and Challenges}
\label{sec:FutureWorks}

Several promising research directions could advance personalized Federated Learning (pFL):

1. Convergence Analysis of pFL on Saddle Point Functions: Investigate how personalized strategies impact the convergence speed and stability of pFL algorithms with saddle point functions, building on ~\cite{ghosh2021escaping} and ~\cite{vlaski2020second}.

2. Incentive Mechanism Design for Personalized FL: Develop mechanisms to encourage participation in decentralized pFL, balancing incentives with privacy and security, as explored by ~\cite{khan2023pi}'s token-based system. Research should assess the impact on engagement and performance.

3. Personalization Strategies in Federated Bandits: Incorporate personalized strategies into federated bandits, such as linear contextual bandits, to improve exploration and exploitation efficiency across diverse client contexts ~\cite{shi2021federated, huang2021federated}.

4. Hybrid Approaches for pFL: Explore combining pFL with transfer learning, meta-learning, or reinforcement learning to enhance model adaptation and generalization across applications.

5. Real-World Dataset Creation and Evaluation: Develop diverse, large-scale datasets reflecting real-world heterogeneity for evaluating privacy-preserving pFL, especially for textual data under multiple attacks.
\section{Conclusion}
\label{sec:Conclusion}

In conclusion, this paper presents a comprehensive analysis of pFL techniques, addressing key challenges such as non-i.i.d data and model performance. By evaluating ten pFL methods across various metrics—including convergence speed, memory overhead, and accuracy—our study offers a nuanced understanding of their practical implications. Our analysis highlights significant trends and trade-offs, providing valuable insights for selecting the most suitable pFL algorithms for different applications, ultimately guiding future research and practical implementations in fields like healthcare and finance.

\bibliographystyle{IEEEtran}
\bibliography{bibliography.bib} 

\end{document}